\documentclass[journal]{IEEEtran}
\usepackage{amsmath,amssymb,amsfonts}
\usepackage{algorithmic}
\usepackage{array}
\usepackage[caption=false,font=normalsize,labelfont=sf,textfont=sf]{subfig}
\usepackage{textcomp}
\usepackage{stfloats}
\usepackage{cellspace}
\usepackage{cite}
\usepackage{verbatim}
\usepackage{graphicx}
\usepackage{color}
\usepackage{bm}
\usepackage{multirow}
\usepackage{multicol}
\usepackage{color}
\usepackage{soul}
\usepackage[hyphens]{url}
\usepackage[varg]{newtxmath}
\usepackage{enumitem}
\usepackage{tabularx}
\usepackage{here}
\usepackage{enumitem}
\setlist[description]{style=nextline, font=\bfseries}

\def\BibTeX{{\rm B\kern-.05em{\sc i\kern-.025em b}\kern-.08em
    T\kern-.1667em\lower.7ex\hbox{E}\kern-.125emX}}
\usepackage{balance}
\begin{document}
\title{Difficulty-Controllable Multiple-Choice Question Generation Using Large Language Models and Direct Preference Optimization}
\author{Yuto Tomikawa and Masaki Uto
    \thanks{This work has been submitted to IEEE Transactions on Learning Technologies for possible publication. This work was supported by JSPS KAKENHI Grant Numbers 23K20727, 23K17585, 24H00739, 25K00833.

        The authors are with the Graduate School of Informatics and Engineering, University of Electro-Communications, Chofu, Tokyo 182-8585, Japan (e-mail: tomikawa@ai.lab.uec.ac.jp; uto@ai.lab.uec.ac.jp).}}

\markboth{\ \ }
{Difficulty-Controllable Multiple-Choice Question Generation Using Large Language Models and Direct Preference Optimization}

\maketitle
\begin{abstract}
    Difficulty-controllable question generation for reading comprehension has gained significant attention in the field of education as a fundamental tool for adaptive learning support.
    Although several neural question generation methods have recently succeeded in controlling difficulty, conventional approaches still face two major limitations.
    First, they cannot directly generate multiple-choice questions, which are the most widely used question type in educational contexts.
    Second, they are not explicitly trained to optimize the accuracy of difficulty control, leaving room for further improvement in difficulty controllability.
    To address these limitations, this study proposes a novel difficulty-controllable multiple-choice question generation method for reading comprehension which leverages a large language model trained using a direct preference optimization technique to improve the accuracy of difficulty control.
\end{abstract}

\begin{IEEEkeywords}
    Automated Question Generation for Reading Comprehension,
    Item Response Theory,
    Deep Neural Networks,
    Adaptive Learning,
    Adaptive Testing,
    Natural Language Processing.
\end{IEEEkeywords}

\section{Introduction} \label{sec:introduction}

\IEEEPARstart{A}{dvances} in information technology and the rise of generative artificial intelligence (AI) have made it easier for learners to access a wide range of information of varying quality, highlighting the importance of developing reading comprehension skills to accurately identify important information and understand its content.
One common approach to fostering reading comprehension skills is to provide learners with a variety of reading passages and related comprehension questions~\cite{kurdi2020systematic,le2014automatic,rathod2022educational,zhang2021review}.
However, manually creating diverse reading comprehension questions for various reading passages presents a significant challenge due to the extensive time and workload required.
To address this challenge, automatic question generation (QG), in which questions are generated from reading passages using AI techniques, has attracted significant attention~\cite{du2017learning,goto2024enhancing,tomikawa2024adaptive,uto2023difficulty,zhou2018neural}.

Early QG methods were typically based on rule-based or template-based approaches, relying on manually crafted templates and rules to generate questions~\cite{kurdi2020systematic,lindberg2013generating,mitkov2003computer}.
However, creating templates or rules capable of generating diverse, high-quality questions incurs considerable costs.
To overcome this issue, QG methods based on deep neural networks that do not require templates or rules have been proposed in recent years~\cite{dutulescu2024beyond,maity2024novel,shuai2023qdg,leite2023towards,li2024planning}.
While early studies on neural QG have primarily employed recurrent neural networks (RNNs), recent QG methods have used transformer models~\cite{vaswani2017attention}.
In particular, numerous approaches using pre-trained language models (PLMs)~\cite{devlin2019bert,lewis2019bart,radford2019language,raffel2020exploring} and their extension into large language models (LLMs)~\cite{brown2020language,ouyang2022training,touvron2023llama,touvron2023llama2,gemini2023gemini,openai2023gpt,rafailov2024direct,jiang2024mixtral,abdin2024phi,yang2024qwen2}, both built on transformer architectures, have been proposed~\cite{lee2024few,liang2023prompting,lin2024prompting,maity2024novel}.

While these methods enable the generation of high-quality questions, they lack the ability to control the difficulty of the generated questions.
When considering the use of QG in the context of learning support and educational assessment, it is desirable to generate questions with controllable difficulty levels.
For example, difficulty control in QG could facilitate adaptive question presentation tailored to the ability level of individual learners, thereby enhancing adaptive learning systems~\cite{ueno2017irt}.
Furthermore, it would be useful for expanding item pools required for managing large-scale examinations~\cite{fuchimoto2022hybrid}.

\begin{figure*}[t]
    \centering
    \includegraphics[width=1.00\textwidth,keepaspectratio=true]{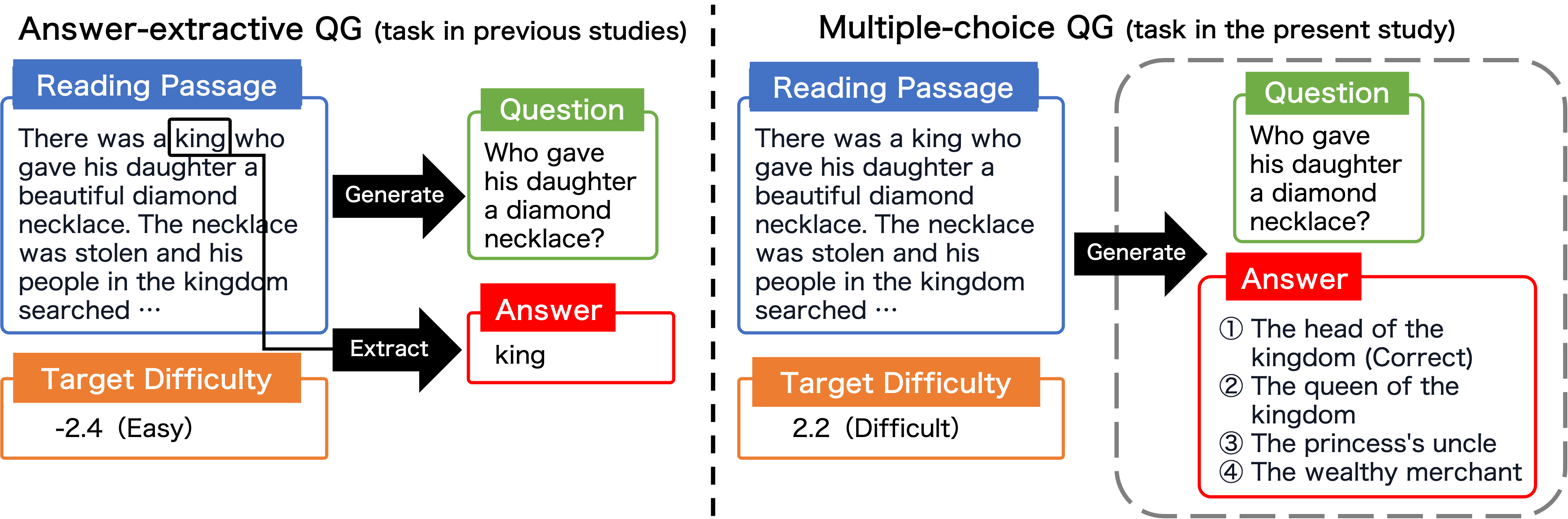}
    \caption{Conceptual diagram of answer-extractive QG (task in previous studies~\cite{tomikawa2024adaptive}) and multiple-choice QG (task in the present study).}
    \label{fig:concept}
\end{figure*}

For these reasons, several difficulty-controllable QG methods have been proposed in recent years~\cite{cheng2021guiding,gao2018difficulty,goto2024enhancing,tomikawa2024difficulty,uto2023difficulty,li2024planning}.
For example, Gao et al.~\cite{gao2018difficulty} proposed a method that generates questions with two levels of difficulty: ``easy'' and ``difficult''.
In addition, Cheng et al.~\cite{cheng2021guiding} proposed a difficulty-controllable QG method that considers the number of sentences that must be referenced in order to reach an answer as the measure of difficulty.
However, Tomikawa et al.~\cite{tomikawa2024adaptive} pointed out that these methods cannot generate questions with difficulty levels tailored to individual learners' abilities, and proposed a method that utilizes item response theory (IRT)~\cite{lord2012applications}, which enables the association between learners' abilities and question difficulty to be interpreted.
Specifically, they developed a method to generate questions across difficulty levels quantified by IRT, proposing a framework for adaptive QG that iteratively estimates learners' abilities and generates questions with difficulty levels appropriate to those abilities.

Although conventional methods have successfully controlled difficulty, they still face two major limitations.
\begin{enumerate}
    \item These methods aim to generate questions with extractive answers, a question type that requires identifying word spans as answers from a given reading passage, as shown on the left side of Figure~\ref{fig:concept}.
          Consequently, they cannot directly generate multiple-choice questions, which are widely used in educational settings.
    \item These methods typically train neural QG models by maximizing the conditional likelihood of each question text given a difficulty level and a reading passage.
          However, they do not explicitly optimize the accuracy of difficulty control, leaving room for further improvement in controllability.
\end{enumerate}

To overcome these limitations, in this study, we develop a difficulty-controllable multiple-choice QG method using an LLM with a training approach that explicitly optimizes difficulty control accuracy through direct preference optimization (DPO)~\cite{rafailov2024direct}.
We first propose a method for generating both question texts and options for multiple-choice questions from a reading passage and a specified difficulty level, as shown on the right side of Figure~\ref{fig:concept}.
This method is implemented using Llama 3.1~\cite{dubey2024llama3}, an open-source LLM.
Next, we introduce a method for maximizing difficulty control accuracy based on DPO, a technique derived from reinforcement learning from human feedback (RLHF)~\cite{ouyang2022training}, which aligns the output of LLMs with specific objectives.
While conventional methods are trained with supervised fine-tuning (SFT) to maximize the conditional likelihood of each question text given a difficulty level, the proposed method directly optimizes difficulty control accuracy after conventional SFT training via the DPO approach, thereby improving the control accuracy.
The proposed method requires a dataset consisting of quadruples of (reading passage, question text, options, question difficulty) for model training, although typical multiple-choice question datasets do not include difficulty levels.
To address this, we employ a method for constructing a multiple-choice question dataset incorporating difficulty levels, following the difficulty estimation method proposed by Tomikawa et al.~\cite{tomikawa2024adaptive}, which uses IRT~\cite{linden2017handbook,lord2012applications,rasch1993probabilistic} and question answering (QA) systems.

This study evaluates the effectiveness of the proposed method through experiments using the RACE dataset~\cite{lai2017race}, which is widely used in the task of automatic multiple-choice reading comprehension QG.

\section{Related Works}
In this section, we provide an overview of conventional neural QG methods and difficulty-controllable QG methods.

\subsection{Neural QG for Reading Comprehension}
Early neural QG methods for reading comprehension have primarily used RNN-based models~\cite{du2017learning,subramanian2018neural,zhou2018neural}.
For example, Du et al.~\cite{du2017learning} proposed a method that generates question text by inputting a reading passage and answer into an RNN encoder, with the obtained output feature vector subsequently fed into an RNN decoder to sequentially generate the question text.
Zhou et al.~\cite{zhou2018neural} proposed an extended RNN-based QG method that incorporates part-of-speech tags for every word to represent word classes.

In contrast, transformer-based models have recently demonstrated superior performance compared with RNN-based models across various natural language processing tasks~\cite{devlin2019bert,dubey2024llama3,lewis2019bart,radford2019language,uto2020neural}.
Therefore, numerous QG methods based on transformers have been proposed in recent years~\cite{durmus2020feqa,lee2022type,lewis2019bart}.
For example, several approaches to QG have been proposed that utilize transformer-based PLMs~\cite{lee2022type,durmus2020feqa,goto2024enhancing,tomikawa2024difficulty} such as T5 (text-to-text transfer transformer)~\cite{raffel2020exploring} and BART (bidirectional and auto-regressive transformer)~\cite{lewis2019bart}.
Moreover, recent studies have proposed QG methods that utilize in-context learning with transformer-based LLMs~\cite{brown2020language,lee2024few,wei2022chain,liang2023prompting,lin2024prompting}.

In addition, several QG methods focusing on multiple-choice questions using PLMs and LLMs have also been proposed~\cite{dutulescu2024beyond,maity2024novel,shuai2023qdg}.
For example, Shuai et al.~\cite{shuai2023qdg} proposed a multiple-choice QG method that combines PLMs with graph neural networks, while Dutulescu et al.~\cite{dutulescu2024beyond} introduced a method that integrates PLMs with knowledge bases for generating multiple-choice questions.
Furthermore, Maity et al.~\cite{maity2024novel} proposed a method for generating multiple-choice questions, using GPT-4~\cite{openai2023gpt} and decomposing the generation process into several sequential subtasks.

Although these methods achieve high-quality QG, they cannot control the difficulty of the generated questions.
As discussed in Section~\ref{sec:introduction}, such control would be highly desirable for educational applications.

\subsection{Difficulty-Controllable Neural QG for Reading Comprehension}

For these reasons, several approaches for difficulty-controllable QG have been proposed in recent years~\cite{cheng2021guiding,gao2018difficulty,goto2024enhancing,tomikawa2024adaptive,tomikawa2024difficulty,uto2023difficulty}.
For example, Gao et al.~\cite{gao2018difficulty} proposed a QG method with two-level difficulty control.
In their study, questions correctly answered by two QA systems were labeled as ``easy,'' while those incorrectly answered by both were labeled as ``difficult,'' with questions yielding mixed results being excluded.
An RNN-based model was then trained to generate a question from a reading passage, answer, and difficulty label as inputs.
Furthermore, Cheng et al.~\cite{cheng2021guiding} proposed a difficulty-controllable QG method that defines difficulty as the number of sentences that must be referenced in order to reach the answer.
In this method, first, a knowledge graph rooted in the answer is constructed from the reading passage.
Next, the reading passage, the answer, and the knowledge graph are utilized to generate a question that can be solved from a specific sentence in the passage.
Finally, based on the knowledge graph, the generated question is iteratively revised until the number of sentences required for reference reaches the specified value, thereby controlling the difficulty.

However, the difficulty levels used in these methods cannot be interpreted in relation to learner ability, making it impossible to generate questions tailored to learner ability levels.
To address this issue, Tomikawa et al.~\cite{tomikawa2024adaptive} proposed a method that uses IRT, a mathematical model that enables the interpretation of difficulty levels in relation to learner ability.
Specifically, they developed a method capable of generating questions aligned with difficulty levels quantified by IRT and proposed a framework for adaptive QG, which iteratively estimates learner ability and generates questions with difficulty levels appropriate to those abilities.

In the present study, we aim to build upon their method.
Thus, we provide a detailed explanation of their method in the next section.

\subsection{Difficulty-Controllable QG Using Item Response Theory} \label{sec:related-irt-difficulty-qg}
Tomikawa et al.~\cite{tomikawa2024adaptive} aimed to generate a question $\bm{q}$ and its corresponding answer $\bm{a}$ from a reading passage $\bm{c}$ and a desired difficulty level $b$,
where the answer $\bm{a}$ exists within the reading passage $\bm{c}$, namely, $\bm{a} \in \bm{c}$, as shown on the left side of Figure~\ref{fig:concept}.

Constructing a QG model for this task requires a dataset consisting of these data, represented as $\{\bm{c}_i, \bm{q}_i, \bm{a}_i, b_i \mid i \in \{1 \ldots I\}\}$, where $\bm{c}_i$, $\bm{q}_i$, $\bm{a}_i$, and $b_i$ represent the $i$-th reading passage, question, answer, and difficulty level, respectively, and $I$ denotes the total number of data instances.
However, general datasets do not include the difficulty levels of questions.
To address this, Tomikawa et al.~\cite{tomikawa2024adaptive} proposed a method for estimating the difficulty of each question in the dataset based on IRT and QA systems.

In IRT, the probability of a learner answering a question correctly is formulated using parameters that represent learner ability and the characteristics of the question, such as its difficulty.
This enables learner ability and question characteristics to be estimated separately.
The Rasch model~\cite{rasch1993probabilistic} was used as the IRT model, which is defined by the following equation:
\begin{equation}
    \label{eq:rasch}
    P_i(\theta_m) = \frac{1}{1 + \exp{(-(\theta_m - b_i))}}
\end{equation}
where $P_i(\theta_m)$ represents the probability that learner $m$ correctly answers question $i$, $\theta_m$ represents the ability of learner $m$, and $b_i$ represents the difficulty level of question $i$.
These model parameters are estimated using maximum likelihood estimation or Bayesian estimation from a collection of response data~\cite{uto2023bayesian,uto2020generalized}.

One of the advantages of utilizing IRT is that it enables the selection of an optimal difficulty level for a learner with a specific ability $\theta$, according to the desired probability of a correct response.
For example, selecting questions whose difficulty corresponds to a learner's ability can be regarded as a reasonable strategy for difficulty selection because presenting questions that learners can answer correctly with a probability of 50\% is known to be effective for enhancing their ability~\cite{ueno2017irt}, and also because the Rasch model satisfies $P_i(\theta_m) = 0.5$ when $b_i = \theta_m$.

The procedure for constructing a question dataset with difficulty levels estimated based on the Rasch model is as follows.
First, each question $\bm{q}_i$ in the dataset is presented to multiple QA systems with varying performance levels, and their correct/incorrect response data are used to estimate the difficulty $b_i$ of each question based on the Rasch model.
Although response data from human learners ideally need to be used, this approach is substituted with QA systems as virtual learners because of the substantial cost of obtaining human data.
Thus, by adding the estimated $b_i$ to the original dataset, a question dataset with difficulty levels is constructed.

Using this dataset, the following two models are trained to enable the difficulty-controllable generations of answers and questions.
\begin{enumerate}
    \item \textbf{Difficulty-Controllable Answer-Extraction Model:}
          This is a BERT (bidirectional encoder representations from transformers)~\cite{devlin2019bert} model designed to extract answers from the reading passage based on the desired difficulty level.
          Specifically, it takes a concatenated string of difficulty level $b$ and the reading passage $\bm{c}$ as input and outputs the start and end positions of the answer $\bm{a}$ within the reading passage.
    \item \textbf{Difficulty-Controllable Answer-Aware Question Generation Model:}
          This is a T5~\cite{raffel2020exploring} model designed to generate questions based on the reading passage, answer, and desired difficulty level.
          Specifically, it takes a concatenated string of difficulty level $b$, reading passage $\bm{c}$, and answer $\bm{a}$ as input and generates the question text $\bm{q}$.
\end{enumerate}

Although this method successfully achieved QG with difficulty control, it still faces two major limitations, as described in Section~\ref{sec:introduction}.
First, it is unable to directly generate multiple-choice questions.
Second, it is not explicitly trained to optimize the accuracy of difficulty control, as detailed in Section~\ref{sec:sft}, leaving room for further improvement in controllability.
To overcome these limitations, we propose a difficulty-controllable QG method for multiple-choice questions using an LLM with a training approach that optimizes difficulty control accuracy through DPO.

\section{Proposed Method}
As shown on the right in Figure~\ref{fig:concept}, we propose a method that takes a reading passage and difficulty level quantified by IRT as input and generates a question, a correct answer option, and distractor options tailored to the specified difficulty level.
Specifically, the proposed QG method is designed as an LLM that receives a prompt $\bm{x}$ consisting of the reading passage $\bm{c}$ and difficulty level $b$ in the input format shown in Table~\ref{tb:prompt}, and outputs the question $\bm{q}$, the correct answer option $\bm{a}$, and the three distractors $\bm{d}_1, \bm{d}_2, \bm{d}_3$ in the output format $\bm{y}$ shown in Table~\ref{tb:prompt}.
We use Llama 3.1~\cite{dubey2024llama3} as the base LLM and train it to incorporate the capability for difficulty control.

\begin{table}[t]
    \caption{Inputs and outputs for the proposed QG model.}
    \label{tb:prompt}
    \resizebox{\columnwidth}{!}{
        \begin{tabular}{p{\columnwidth}}
            \hline
            { \textbf{Input ($\bm{x}$):}}                                                                                                                                                                                                                                                                                                                                                                                                                                                                                                      \\
            \#\#\# Instruction: Create a question and four options with a difficulty level of $\{b\}$ based on the Context. Option 1 is the correct answer and Options 2, 3, and 4 are distractor options. The difficulty level $-3.0$ is the easiest and 3.0 is the most difficult. The output format is ``$\langle$c$\rangle$ Option 1 (Correct Option) $\langle$q$\rangle$ Question $\langle$d1$\rangle$ Option 2 (Distractor Option) $\langle$d2$\rangle$ Option 3 (Distractor Option) $\langle$d3$\rangle$ Option 4 (Distractor Option)'' \\\\
            \#\#\# Context: \{$\bm{c}$\}                                                                                                                                                                                                                                                                                                                                                                                                                                                                                                       \\\\
            \#\#\# Response:                                                                                                                                                                                                                                                                                                                                                                                                                                                                                                                   \\
            \hline
            { \textbf{Output ($\bm{y}$):}}                                                                                                                                                                                                                                                                                                                                                                                                                                                                                                     \\
            { $\langle$c$\rangle$ \{$\bm{a}$\} $\langle$q$\rangle$ \{$\bm{q}$\} $\langle$d1$\rangle$ \{$\bm{d}_1$\} $\langle$d2$\rangle$ \{$\bm{d}_2$\} $\langle$d3$\rangle$ \{$\bm{d}_3$\}}                                                                                                                                                                                                                                                                                                                                                   \\\hline
        \end{tabular}
    }
\end{table}

\subsection{Constructing a Multiple-Choice Question Dataset with Difficulty Levels} \label{sec:dataset}

Our method requires a dataset $\mathcal{D} = \{\bm{c}_i, \bm{q}_i, \bm{a}_i, \bm{d}_{i1}, \bm{d}_{i2}, \bm{d}_{i3}, b_i | i \in \{1 \ldots I\}\}$, where $\bm{a}_i$, $\bm{d}_{i1}$, $\bm{d}_{i2}$, and $\bm{d}_{i3}$ represent the correct option and three distractors for $i$-th question $\bm{q}_i$, all of which are related to the $i$-th reading passage $\bm{c}_i$.
However, because general datasets for QG do not include the difficulty level $b$, we construct a dataset with difficulty levels following the procedure of Tomikawa et al.~\cite{tomikawa2024adaptive}.
The construction of this dataset is as follows:
\begin{enumerate}
    \item We construct multiple QA systems that predict the correct answer option given a reading passage, a question, and four options.
          For the detailed design of the QA system, see~\cite{liu2019roberta}.
          To ensure diverse ability levels among QA systems, we construct them based on various pre-trained transformer models by varying the amount of training data, as detailed in Section~\ref{sec:eval-process}.
    \item Each question in the dataset is presented to the constructed QA systems, and the response data are used to estimate the difficulty level of each question based on the Rasch model.
    \item The estimated difficulty values are added to the original dataset, thereby constructing a dataset that includes difficulty levels.
\end{enumerate}

\subsection{Initial Fine-tuning} \label{sec:sft}

Using the constructed dataset, the proposed method initially trains Llama 3.1 by minimizing the following negative log-likelihood as the loss function:
\begin{equation}
    \mathcal{L}_{\mathrm{sft}} = -\mathbb{E}_{(\bm{x}, \bm{y}) \sim \mathcal{D}} \left [ \log \pi(\bm{y} | \bm{x}, \bm{\phi}_{\mathrm{sft}})\right ]
\end{equation}
where $\pi(\bm{y}|\bm{x}, \bm{\phi}_{\mathrm{sft}})$ represents the likelihood of the output string $\bm{y}$ conditioned on the input $\bm{x}$ calculated based on Llama 3.1, and $\bm{\phi}_{\mathrm{sft}}$ represents the model parameters to be trained.
The training procedure follows that of the conventional difficulty-controllable QG model proposed by Tomikawa et al.~\cite{tomikawa2024adaptive}, with the exception that Llama 3.1 is used as the base LLM instead of T5, and the detailed design of the input and output formats is different.

\subsection{Direct Preference Optimization for Improving Difficulty Control Accuracy} \label{sec:dpo}
\begin{figure*}[h]
    \centering
    \includegraphics[width=0.75\textwidth,keepaspectratio=true]{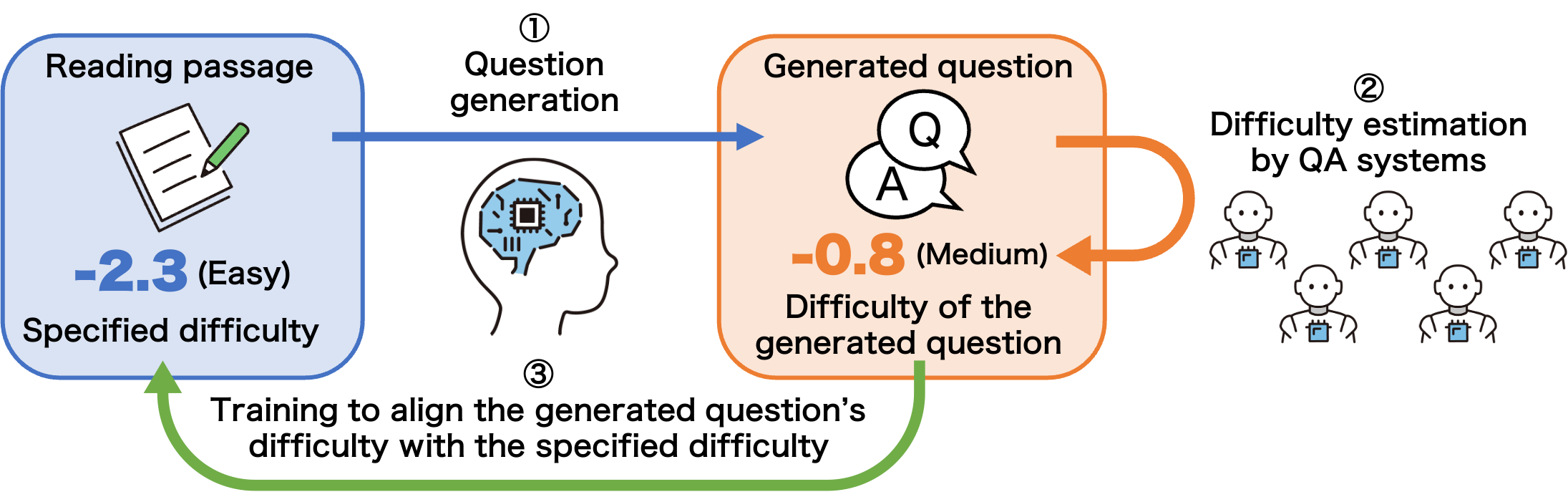}
    \caption{Outline of RLHF for optimizing the accuracy of difficulty control.}
    \label{fig:rlhf-concept}
\end{figure*}

By performing the initial training, Llama 3.1 becomes capable of difficulty-controllable generation of questions and options in the format of $\bm{y}$.
However, this training does not directly optimize difficulty control accuracy as the objective function.
Therefore, we further propose a method to explicitly optimize difficulty control accuracy as the objective function.
For this training, we use DPO~\cite{rafailov2024direct}, which is a supervised learning method proposed to efficiently compute RLHF~\cite{ouyang2022training} and has been widely used as an additional training approach for recent LLMs~\cite{dubey2024llama3,jiang2024mixtral}.

RLHF is a reinforcement learning method designed to align the output of an LLM with specific preferences based on human feedback.
One of its advantages is its ability to optimize arbitrary functions, including those that are non-differentiable.
When using RLHF to optimize the accuracy of difficulty control in our context, it is reasonable to train the model to align the difficulty of the generated questions with the specified difficulty, as shown in Figure~\ref{fig:rlhf-concept}.
More specifically, the objective function to be maximized can be defined as follows:
\begin{align} \label{eq:rlhf}
    \begin{split}
        \fontsize{9.5pt}{12pt}\selectfont
        J(\bm{\phi}_\mathit{rlhf}) & = \mathbb{E}_{\bm{x} \sim \mathcal{D}, \bm{y}' \sim \pi(\bm{y}' | \bm{x}, \bm{\phi}_\mathit{rlhf})} \left [ r(\bm{x}, \bm{y}') \right ] -             \\
                                   & \quad \beta \cdot \mathit{KL} \left ( \pi(\bm{y}' | \bm{x}, \bm{\phi}_\mathit{rlhf}) || \pi (\bm{y}' | \bm{x}, \hat{\bm{\phi}}_\mathrm{sft}) \right )
    \end{split}
\end{align}
where $\bm{y}'$ represents a question generated from the input $\bm{x}$ by the model being trained, $\beta$ is a hyperparameter.
Furthermore, $\mathit{KL} \left ( \pi(\bm{y}' | \bm{x}, \bm{\phi}_\mathit{rlhf}) || \pi (\bm{y}' | \bm{x}, \hat{\bm{\phi}}_\mathrm{sft}) \right )$ represent the Kullback-Leibler divergence of $\pi(\bm{y}' | \bm{x}, \bm{\phi}_\mathit{rlhf})$ from $\pi (\bm{y}' | \bm{x}, \hat{\bm{\phi}}_\mathrm{sft})$, wherein $\hat{\bm{\phi}}_\mathrm{sft}$ represents the parameters of the model constructed through the initial fine-tuning in Section~\ref{sec:sft} and $\bm{\phi}_\mathit{rlhf}$ represents the model parameters to be trained, with its initial values set to $\hat{\bm{\phi}}_\mathrm{sft}$.
Moreover, $r(\bm{x}, \bm{y}')$ is a reward function representing difficulty control accuracy, defined as the negative squared error between the specified difficulty in the input $\bm{x}$ and the difficulty of the generated question $\bm{y}'$.
The difficulty of the generated question $\bm{y}'$ can be estimated using the Rasch model, based on responses from the multiple QA systems given their ability estimates $\theta$.

However, training with RLHF presents challenges, such as a tendency for unstable convergence and high computational costs due to the need to repeatedly generate questions during the parameter update process to calculate the reward function~\cite{rafailov2024direct}.
To address these issues, DPO, a supervised learning method mathematically equivalent to RLHF, has been proposed.
The loss function for DPO corresponding to the above RLHF can be derived as follows:
\begin{equation} \label{eq:dpo}
    \begin{split}
        \mathcal{L}_{\mathrm{dpo}} = -\mathbb{E}_{(\bm{x}, \bm{y}_w, \bm{y}_l) \sim \mathcal{D}_{\mathrm{dpo}}} \left[ \log \sigma \left( \beta \log \frac{\pi (\bm{y}_w | \bm{x}, \bm{\phi}_{\mathrm{dpo}})}{\pi (\bm{y}_w | \bm{x}, \hat{\bm{\phi}}_{\mathrm{sft}})} - \right.\right. \\
            \left. \left. \beta \log \frac{\pi (\bm{y}_l | \bm{x}, \bm{\phi}_{\mathrm{dpo}})}{\pi (\bm{y}_l | \bm{x}, \hat{\bm{\phi}}_{\mathrm{sft}})} \right) \right]
    \end{split}
\end{equation}
where $\sigma$ represents the sigmoid function; $\bm{\phi}_\mathrm{dpo}$ represents the model parameters to be trained, initialized with $\hat{\bm{\phi}}_\mathrm{sft}$ as the starting parameters;
$\bm{y}_w$ and $\bm{y}_l$ represent questions satisfying $r(\bm{x}, \bm{y}_w) > r(\bm{x}, \bm{y}_l)$, meaning that $\bm{y}_w$ has higher difficulty control accuracy compared to $\bm{y}_l$;
and $\mathcal{D}_{\mathrm{dpo}}$ represents the training dataset for DPO, defined as $\mathcal{D}_{\mathrm{dpo}} = \{\bm{x}_i, \bm{y}_{wi}, \bm{y}_{li} \mid i \in \{1 \ldots I\}\}$.

In RLHF, it is necessary to repeatedly generate questions $\bm{y}'$ from the trained QG model and evaluate their difficulty throughout the training process.
In contrast, DPO can solve the same optimization solely using the original training dataset, $\mathcal{D}$, by transforming it into the format $\mathcal{D}_{\mathrm{dpo}}$.
Specifically, the procedure for constructing $\mathcal{D}_{\mathrm{dpo}}$ is as follows:
\begin{enumerate}
    \item For input $\bm{x}_i$ consisting of reading passage $\bm{c}_i$ and difficulty level $b_i$, we construct output $\bm{y}_i$ consisting of the corresponding question text $\bm{q}_i$, correct answer $\bm{a}_i$, and distractors $(\bm{d}_{i1}, \bm{d}_{i2}, \bm{d}_{i3})$ and treat it as the good output $\bm{y}_{wi}$.
    \item For the same input $\bm{x}_i$, bad output $\bm{y}_{li}$ is constructed by randomly selecting another question that corresponds to the same reading passage used in $\bm{x}_i$.
\end{enumerate}
Applying the above procedure to all data, $i \in \{1, \ldots, I\}$, we can construct the training dataset $\mathcal{D}_{\mathrm{dpo}}$.
Note that DPO assumes that there are multiple questions related to each reading passage in the dataset, a requirement that typical datasets generally meet.
Using this dataset, the proposed method can directly optimize difficulty control accuracy by minimizing Eq.~(\ref{eq:dpo}) as the loss function.

\section{Experiments}
In this section, we describe the evaluation experiments conducted to assess the effectiveness of the proposed method.

\subsection{Experimental Procedures} \label{sec:eval-process}

Our experiments used the RACE dataset~\cite{lai2017race}, which has been widely used for model training and performance evaluation in a variety of studies on multiple-choice reading comprehension QG.
The RACE dataset consists of multiple-choice reading comprehension questions created by English teachers, including 87866 training instances, 4887 validation instances, and 4934 test instances.
In our experiments, we used the RACE training dataset to train the QG model and the RACE validation dataset to construct the QA systems.
Hereafter, we denote the RACE training dataset as the QG model training dataset $\mathcal{D}^{(\mathrm{qg})}$ and the RACE validation dataset as the QA system training dataset $\mathcal{D}^{(\mathrm{qa})}$.

Each subset of the RACE dataset consists of a reading passage $\bm{c}$, an associated question $\bm{q}$, the corresponding correct answer option $\bm{a}$, and three distractor options $\bm{d}_1$, $\bm{d}_2$ and $\bm{d}_3$.
More specifically, each data instance can be represented as $\{\bm{c}_i, \bm{q}_i, \bm{a}_i, \bm{d}_{i1}, \bm{d}_{i2}, \bm{d}_{i3} \mid i \in \{1, \ldots, I\}\}$.
Moreover, this dataset includes multiple questions for each reading passage, which meets our requirement for DPO training.
However, because the RACE dataset does not include difficulty levels, our experiments begin with the construction of an extended RACE dataset incorporating IRT-based difficulty following the procedures in Section~\ref{sec:dataset}.

The specific experimental procedure, including dataset construction, was performed as follows:

\begin{description}
    \item[\textbf{1. Construction of QA Systems}:]
          Using the QA system training dataset $\mathcal{D}^{(\mathrm{qa})}$, we constructed various QA systems with various levels of ability.
          Specifically, the 10 pre-trained transformer models listed below were configured as multiple-choice QA systems, which take a reading passage, a question, and four optionss as input and then predict the correct answer option.
          \begin{quote}\it
              bert-base-uncased, bert-large-uncased, roberta-base, roberta-large, albert-base-v1, albert-base-v2, albert-large-v1, albert-large-v2, deberta-v3-base, deberta-v3-large
          \end{quote}
          These models were subsequently trained on the entire $\mathcal{D}^{(\mathrm{qa})}$.
          Next, for each of the 10 trained models, the parameters of the output layer were randomly initialized and re-trained using randomly selected 10\%, 20\%, $\ldots$, 100\% of $\mathcal{D}^{(\mathrm{qa})}$, while keeping the parameters in other layers fixed.
          Through this procedure, we prepared 100 QA systems with varying levels of ability.
          However, because extremely low-performing QA systems might introduce noise, we filtered out such systems.
          Specifically, considering that a random selection from four options would achieve a 25\% chance rate, only QA systems with an accuracy rate of at least 30\% on the QG model training dataset $\mathcal{D}^{(\mathrm{qg})}$ were retained.
          As a result, 77 QA systems remained.
    \item[\textbf{2. Construction of a QG Model Training Dataset with Difficulty Levels}:]
          Using the 77 QA systems, we collected their correct/incorrect response data for all questions in the QG model training dataset $\mathcal{D}^{(\mathrm{qg})}$.
          Based on these response data, we next estimated the difficulty $b_i$ of each question using the Rasch model and added these values to $\mathcal{D}^{(\mathrm{qg})}$ to construct the QG model training dataset with difficulty levels.
          For the estimation, marginal maximum likelihood estimation~\cite{bock1981marginal} was employed.
          The ability values of each QA system were also estimated along with the difficulty estimation for the subsequent analysis.
          The ability values were estimated using the maximum likelihood estimation, conditioned on the question difficulties.
    \item[\textbf{3. Initial Fine-tuning of the Proposed QG Model}:]
          Using this dataset, we initially fine-tuned the proposed multiple-choice QG model on \texttt{Llama-3.1-8b}\footnote{\url{https://huggingface.co/meta-llama/Llama-3.1-8B}} with QLoRA~\cite{dettmers2023qlora}, using the approach of maximizing the likelihood of the question and its four options, as described in Section~\ref{sec:sft}.
          Hereafter, this fine-tuned model is referred to as $QG_{\mathrm{sft}}$.
    \item[\textbf{4. Applying DPO}:]
          Given the fine-tuned model $QG_{\mathrm{sft}}$, we applied DPO with QLoRA following the procedures described in Section~\ref{sec:dpo}.
          Hereafter, this model is referred to as $QG_{\mathrm{dpo}}$.
    \item[\textbf{5. Question Generation for Various Difficulty Levels}:]
          For the 300 reading passages randomly selected from the RACE test dataset, we specified difficulty levels $b^{(\mathrm{specified})}$ from $-3.0$ to $3.0$ in increments of $0.1$, and constructed a total of 18300 input data instances in accordance with the input format $\bm{x}$ shown in Table~\ref{tb:prompt}.
          Then, we generated questions by inputting these instances into $QG_{\mathrm{sft}}$ and $QG_{\mathrm{dpo}}$, respectively.
          In Sections~\ref{sec:eval-diff} through~\ref{sec:eval-fisher-info} , the quality of the generated questions for each method is evaluated.
\end{description}

\subsection{Evaluation of Accuracy of Difficulty Control} \label{sec:eval-diff}
In our experiments, we first evaluated the accuracy of difficulty control in the proposed method.
More specifically, the difficulty of the questions generated in Procedure~5 was evaluated using the correct and incorrect responses obtained from the 77 QA systems constructed in Procedure~1. The following two approaches were adopted: (1) the correct response ratio, and (2) IRT-based difficulty estimates derived from the Rasch model, conditioned on the ability estimates of the QA systems obtained in Procedure~2.

\begin{figure}[t]
    \centering
    \includegraphics[width=1.0\columnwidth,keepaspectratio=true]{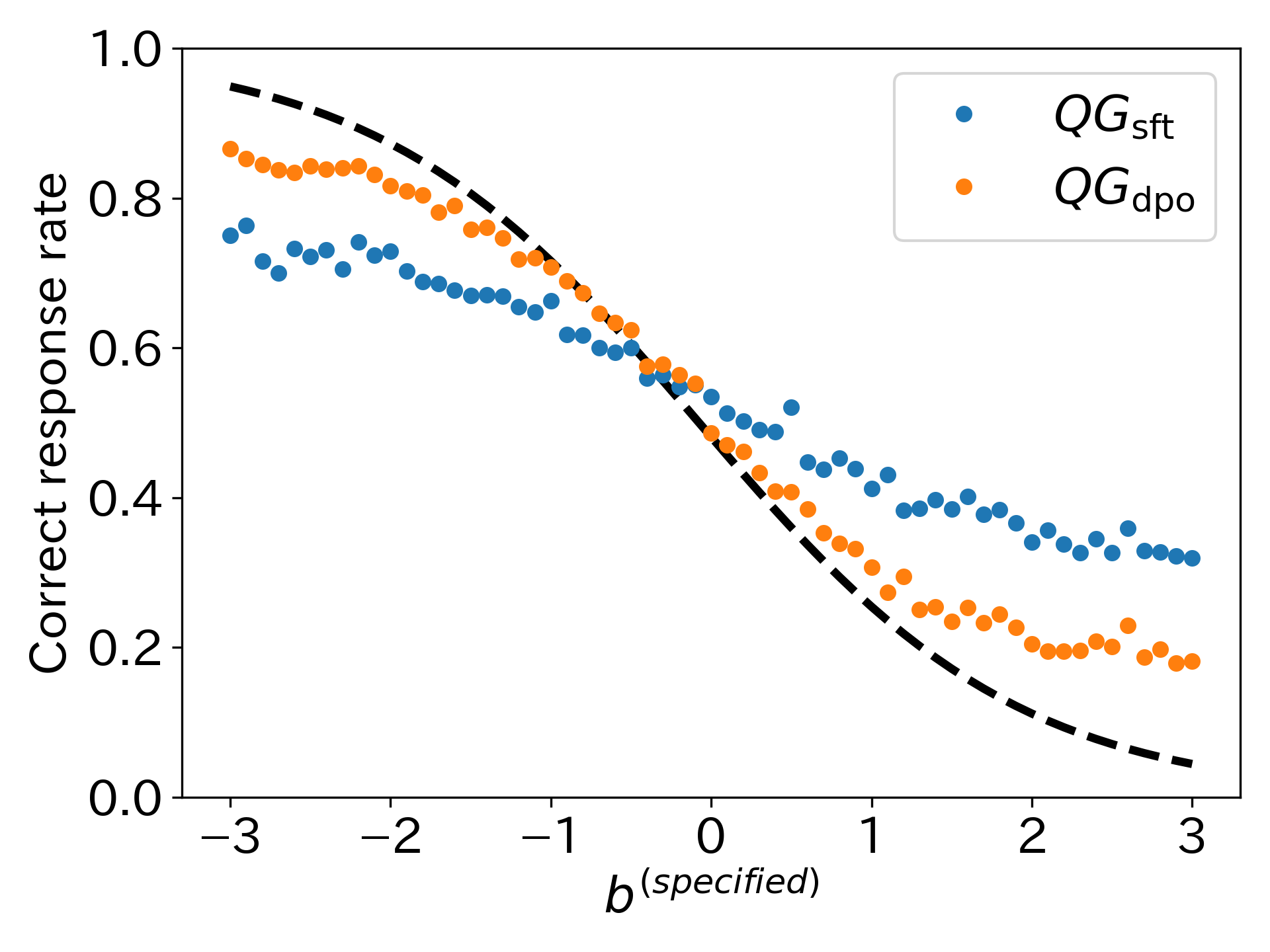}
    \caption{{Relationship between specified difficulty and correct response rates.}}
    \label{fig:correct-ratio}
\end{figure}

Figure~\ref{fig:correct-ratio} shows the results for the correct response rate.
In this figure, the horizontal axis represents the difficulty parameters specified during generation, and the vertical axis represents the correct response rate for the generated questions at each specified difficulty level.
The blue and orange dots represent the rate of correct responses to questions generated by $QG_{\mathrm{sft}}$ and $QG_{\mathrm{dpo}}$, respectively.
The black line represents the correct-response probability for each specified difficulty value, calculated using the Rasch model.
The probabilities were computed based on an ability value of $\theta = -0.08$, which is the median ability of the 77 QA systems.
The black line can be regarded as the theoretical expectation.
The closer the blue and orange plots are to this line, the more accurately the generated questions reflect the specified difficulty level.

\begin{figure}[t]
    \centering
    \includegraphics[width=1.0\columnwidth,keepaspectratio=true]{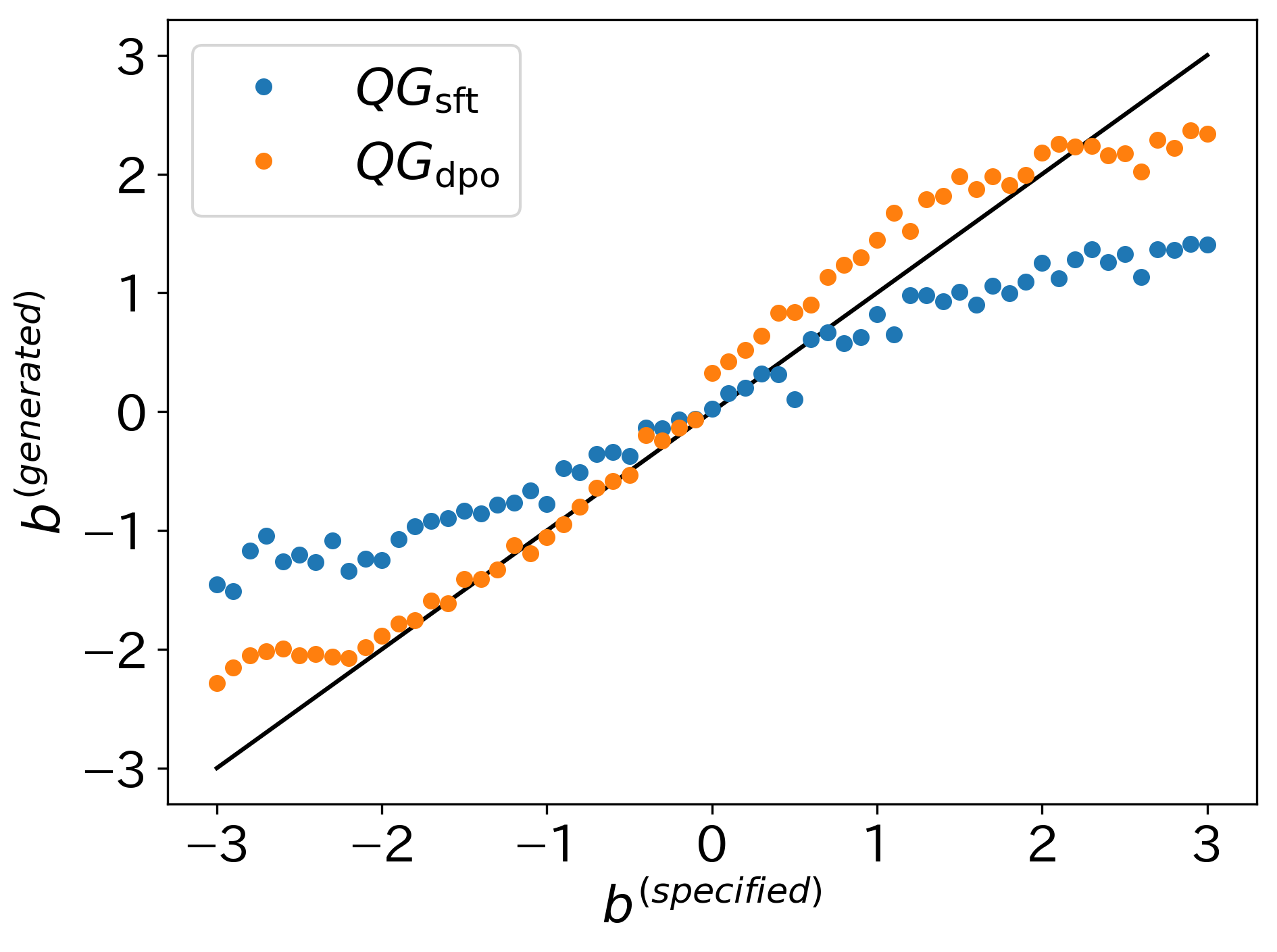}
    \caption{{Relationship between specified and estimated IRT-based difficulty.}}
    \label{fig:b-result}
\end{figure}

Figure~\ref{fig:b-result} shows the relationship between the specified difficulty levels and the IRT-based difficulty estimates $b^{(\mathrm{generated})}$ calculated from the responses of the QA systems.
The horizontal axis represents the specified difficulty levels, while the vertical axis represents the average difficulty estimates for the generated questions at each specified difficulty level.
The blue and orange dots represent the results for $QG_{\mathrm{sft}}$ and $QG_{\mathrm{dpo}}$, respectively.
The black line represents a straight line with a slope of 1 passing through the origin, indicating that the closer the results are to it, the closer the difficulty of the generated questions is to the specified difficulty levels.

These figures show that as the specified difficulty level increases, the correct response rates decrease and the difficulty of the generated questions increase in both $QG_{\mathrm{sft}}$ and $QG_{\mathrm{dpo}}$.
These results indicate that the proposed models successfully achieved difficulty control, regardless of whether DPO was applied.
Furthermore, comparing the two methods makes it evident that $QG_{\mathrm{dpo}}$ exhibits trends closer to the black line in both figures.
For further validation, we also calculated the mean absolute error between the estimated difficulty values and the specified difficulty values, resulting in 1.41 for $QG_{\mathrm{sft}}$ and 1.07 for $QG_{\mathrm{dpo}}$.
These results indicate that incorporating our DPO effectively improves the accuracy of difficulty control.

\subsection{Evaluation based on Fisher Information} \label{sec:eval-fisher-info}

\begin{figure*}[t]
    \includegraphics[width=\textwidth,keepaspectratio=true]{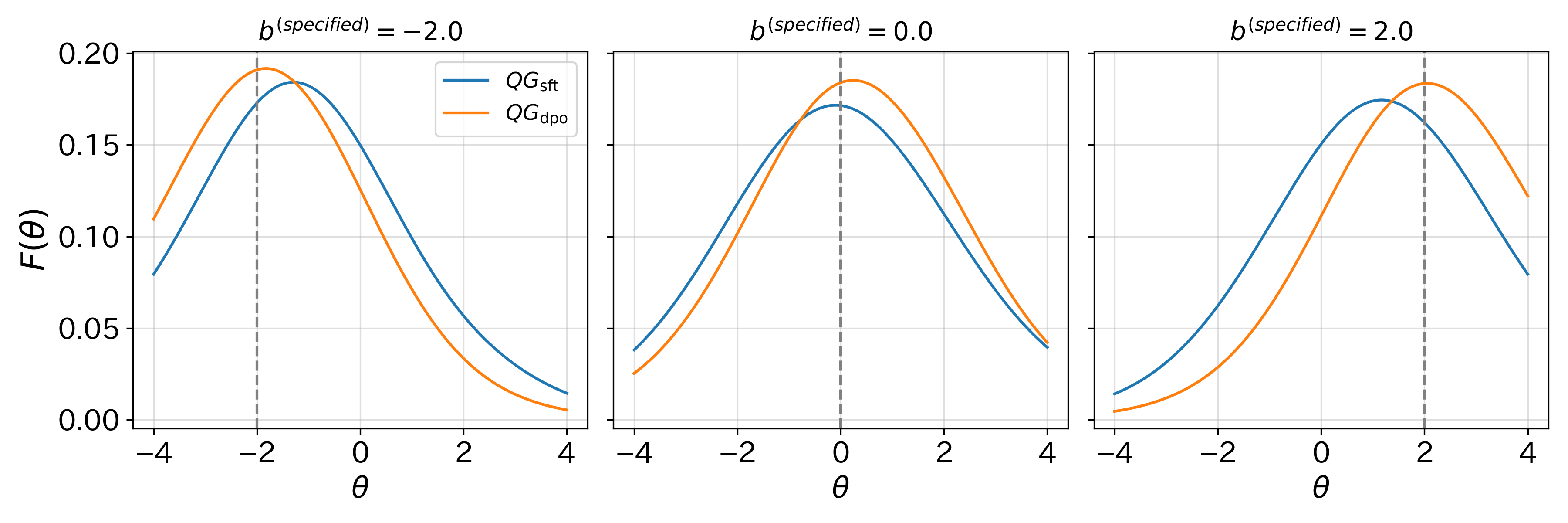}
    \caption{Fisher information curves for questions generated at the specified difficulty $b^{(\text{specified})}=-2.0, 0.0, 2.0$.}
    \label{fig:fisher-information}
\end{figure*}

This section analyzes whether the improvement in difficulty control achieved by the proposed method enhances the accuracy of ability estimation.

IRT makes it possible for the accuracy of ability measurement to be estimated in terms of Fisher information~\cite{lord2012applications}.
The Fisher information of the Rasch model for a question $i$ is defined as a function of learner ability $\theta$ as follows.

\begin{equation}
    \label{eq:fisher}
    F_i(\theta) = P_i(\theta)(1 - P_i(\theta))
\end{equation}
Because questions with higher Fisher information at a specific ability level result in more accurate ability estimation for learners whose abilities are near that level, we investigated whether $QG_{\mathrm{dpo}}$ can generate questions with higher Fisher information compared with $QG_{\mathrm{sft}}$. This investigation involved the experiments conducted as described below.
\begin{enumerate}
    \item We collected the questions generated by $QG_{\mathrm{sft}}$ and $QG_{\mathrm{dpo}}$ through the procedure described in Section~\ref{sec:eval-process}, at specified difficulty levels of $b^{(\mathrm{specified})} = -2.0, 0.0, 2.0$.
    \item We estimated the difficulty parameter for each question, $b^{(\mathrm{generated})}$, in the same manner as described in Section~\ref{sec:eval-diff}.
    \item We computed the Fisher information for each question, given the estimated difficulty $b^{(\mathrm{generated})}$.
\end{enumerate}

The resulting Fisher information curves, averaged within each $b^{(\mathrm{specified})}$ group, are shown in Figure~\ref{fig:fisher-information}.
The horizontal axis shows the ability values $\theta$, and the vertical axis shows the Fisher information, $F(\theta)$.
The blue curve corresponds to questions generated by $QG_{\mathrm{sft}}$, and the orange curve to those generated by $QG_{\mathrm{dpo}}$.
These figures show that $QG_{\mathrm{dpo}}$ yields higher Fisher information near $\theta = b^{(\mathrm{specified})}$ than $QG_{\mathrm{sft}}$.
This suggests that $QG_{\mathrm{dpo}}$ would result in more accurate ability estimation under the adaptive difficulty selection strategy ($b=\theta$) discussed in Section \ref{sec:related-irt-difficulty-qg}.

Beyond the analysis limited to the three specific difficulty levels, we conducted the same experiments over a grid of $b\in [-3.0, 3.0]$ with a step size of 0.1.
The average Fisher information for each specified difficulty $b^{(\mathrm{specified})}$ at the corresponding ability level $\theta = b^{(\mathrm{specified})}$ was 0.186 for $QG_{\mathrm{dpo}}$ and 0.163 for $QG_{\mathrm{sft}}$.
This result also supports the above conclusion.

\subsection{Evaluation of the Fluency, Content Relevance, and Answerability} \label{sec:eval-others}
The above experiments demonstrate that our DPO was effective in improving difficulty control accuracy, although concerns remain regarding the potential degradation of the quality of generated questions.
Therefore, we evaluated the quality of the questions generated by $QG_{\mathrm{sft}}$ and $QG_{\mathrm{dpo}}$ in terms of fluency, content relevance, and answerability.

Previous studies of QG have generally conducted such evaluations through manual assessment. However, this process can be resource-intensive and potentially prone to inconsistency.
Therefore, this study employs LLMs as an alternative to human evaluators, a method known as the {\it LLM-as-a-Judge} approach, which has recently been applied across various text generation evaluation tasks~\cite{abdin2024phi,jiang2024mixtral,moon2024generative,yang2024qwen2,zheng2023judging}.
Use of the LLM-as-a-Judge approach is justified as follows.
First, a previous study applying the LLM-as-a-Judge approach for various natural language evaluation tasks using benchmark datasets (MT-Bench)~\cite{zheng2023judging} has shown a high agreement rate exceeding 80\% between human and LLM evaluations.
Second, a previous QG study~\cite{moon2024generative} also reported a high correlation coefficient of 0.7 between human and LLM evaluations.

Based on the LLM-as-a-Judge approach, we evaluated all the questions generated by $QG_{\mathrm{sft}}$ and $QG_{\mathrm{dpo}}$ in experimental Procedure~5 for specified difficulty levels of $-2.0$, $0.0$, and $2.0$.
The evaluations were conducted based on the following three criteria using GPT-4.1 mini\footnote{\url{https://platform.openai.com/docs/models/gpt-4.1-mini}} as the judge LLM.
\begin{itemize}
    \item \textbf{Fluency:}
          Evaluate the grammar and fluency of the question. Ratings were assigned on a 3-point scale, with 2 meaning appropriate, 1 meaning acceptable, and 0 meaning inappropriate.
    \item \textbf{Content Relevance:}
          Evaluate the relevance of the question to the reading passage.
          Ratings were assigned on a 2-point scale, with 1 meaning appropriate and 0 meaning inappropriate.
    \item \textbf{Answerability:}
          Evaluate the answerability of the question and its options.
          Ratings were assigned on a 2-point scale, with 1 meaning answerable (if there is exactly one correct answer option among the four options) and 0 meaning unanswerable (if there are no correct answer options or more than one correct answer option).
\end{itemize}
The evaluation of fluency and content relevance was conducted in a zero-shot manner using a prompt that included task instructions and explanations of each evaluation criterion.
For answerability, we also leveraged Chain-of-Thought reasoning~\cite{wei2022chain,kojima2022large} by adding an instruction to generate a reasoning process before providing the binary judgement, given that evaluating answerability is relatively complex.
The prompt used for this evaluation is provided in Appendix~\ref{appendix:eval-prompt}.
To compare the quality of the generated questions, we also conducted the same evaluation on human-created questions from the QG model training dataset $\mathcal{D}^{(\mathrm{qg})}$, which had estimated difficulty levels of $-2.0$, $0.0$, and $2.0$.

\begin{table}[t]
    \centering
    \caption{Evaluation of question quality.}
    \label{tb:llm-judge}
    \resizebox{\columnwidth}{!}{
        \begin{tabular}{@{\hspace{0.1cm}}l@{\hspace{0.1cm}}c@{\hspace{0.1cm}}c@{\hspace{0.1cm}}c}
            \hline
                          & RACE Dataset    & $QG_{\mathrm{sft}}$ & $QG_{\mathrm{dpo}}$ \\ \hline
            Fluency       & 1.94 $\pm$ 0.29 & 1.95 $\pm$ 0.26     & 1.94 $\pm$ 0.29     \\
            Relevance     & 0.98 $\pm$ 0.15 & 0.99 $\pm$ 0.12     & 0.97 $\pm$ 0.17     \\
            Answerability & 0.91 $\pm$ 0.29 & 0.72 $\pm$ 0.45     & 0.76 $\pm$ 0.43     \\ \hline
        \end{tabular}
    }
\end{table}

Table~\ref{tb:llm-judge} presents the results, showing the average and standard deviation of the evaluated scores.
The table indicates that the average scores for fluency and content relevance are consistently high and identical across all cases.
The answerability scores for $QG_{\mathrm{sft}}$ and $QG_{\mathrm{dpo}}$ are nearly identical, although they are lower than those of the human-created questions in the $\mathcal{D}^{(\mathrm{qg})}$.

These results suggest that integrating our DPO does not degrade the quality of the generated questions in terms of fluency, content relevance, and answerability, while the possibility of improving answerability remains a common challenge regardless of whether DPO is used.

\subsection{Evaluation of Difficulty Controllability in Few-Shot Learning} \label{sec:eval-fewshot}

LLMs can generally solve various tasks through few-shot learning, an approach in which a few input--output examples are provided in the prompt~\cite{brown2020language}, suggesting the potential feasibility of few-shot difficulty control in QG.
However, the effectiveness of this approach remains uncertain, and thus this section evaluates its feasibility, using GPT-4.1 mini as the base LLM.
The experimental procedure was as follows:
\begin{enumerate}
    \item We randomly selected questions at three difficulty levels, namely $-2.0$, $0.0$, and $2.0$, from the QG model training dataset $\mathcal{D}^{(\mathrm{qg})}$ with difficulty levels to create input--output pairs $(\bm{x}, \bm{y})$.
          Treating a set of $(\bm{x}, \bm{y})$ for three difficulty levels as a one-shot example, we created 3-shot, 10-shot, and 30-shot examples.
    \item A reading passage was randomly selected from the RACE test dataset, and we created the input prompt $\bm{x}$ following Table~\ref{tb:prompt} for each of the three difficulty levels.
    \item These prompts were combined with the few-shot examples and then input into GPT-4.1 mini to generate questions.
    \item By changing the reading passage selected in Procedure 2, we repeated Procedures 2 and 3 a total of 100 times for each difficulty level.
    \item The generated questions were answered by the 77 QA systems constructed in Procedure~1 of Section~\ref{sec:eval-process}.
          The difficulty values of these questions were estimated from the response data using the Rasch model.
\end{enumerate}

Table~\ref{tb:few-shot} shows the results, indicating the average difficulty estimates for the generated questions at each specified difficulty level.
The table also includes the results for questions generated by $QG_{\mathrm{sft}}$ and $QG_{\mathrm{dpo}}$ at the same specified difficulty levels.

\begin{table}[t]
    \caption{ Difficulty controllability results for the proposed method and few-shot learning.}
    \centering
    \label{tb:few-shot}
    \resizebox{\columnwidth}{!}{
        \begin{tabular}{@{\hspace{0.1cm}}c@{\hspace{0.1cm}}c@{\hspace{0.1cm}}c@{\hspace{0.1cm}}c@{\hspace{0.1cm}}c@{\hspace{0.1cm}}c@{\hspace{0.1cm}}c@{\hspace{0.1cm}}c}
            \hline
            \multirow{2}{*}{\begin{tabular}[c]{@{}c@{}}Specified\\ Difficulty\end{tabular}} &  & \multicolumn{2}{c}{Proposed Method} &                     & \multicolumn{3}{c}{Few-Shot Learning}                              \\ \cline{3-4} \cline{6-8}
                                                                                            &  & $QG_{\mathrm{sft}}$                 & $QG_{\mathrm{dpo}}$ &                                       & 3-shot & 10-shot & 30-shot \\ \hline
            -2.0                                                                            &  & -1.25                               & -1.89               &                                       & -1.60  & -1.56   & -1.22   \\
            0.0                                                                             &  & 0.03                                & 0.32                &                                       & -1.71  & -1.55   & -1.55   \\
            2.0                                                                             &  & 1.25                                & 2.18                &                                       & -1.29  & -1.33   & -1.21   \\\hline
        \end{tabular}
    }
\end{table}

As shown in the table, while $QG_{\mathrm{sft}}$ and $QG_{\mathrm{dpo}}$ succeeded in increasing the difficulty of the generated questions as the specified difficulty level increased, the few-shot learning approach revealed no correlation between the generated and specified difficulty, suggesting a failure in difficulty control.
This indicates that few-shot learning is insufficient for achieving difficulty control, further validating the usefulness of the proposed method.

\subsection{Qualitative Analysis of Difficulty Levels}
\label{sec:eval-reasoning-type}

In this section, we examine whether the question difficulties we quantified are related to question complexity.
We use the reasoning type as an index of question complexity because the questions in the RACE dataset are known to be classified into the following four reasoning types~\cite{chen2016thorough,trischler2017newsqa}, ordered from simpler to more complex:
\begin{itemize}
    \item Word matching - questions of this type directly match a text span in the given passage.
    \item Paraphrasing - this type of question is paraphrased or entailed by a single sentence in the passage, and the answer can be extracted from that sentence.
    \item Single-sentence reasoning - questions of this type require inference based on a single sentence in the passage, typically involving recognition of incomplete information or conceptual overlap.
    \item Multi-sentence reasoning - this type of question requires the integration of information distributed across multiple sentences in order to infer the correct answer.
\end{itemize}

\begin{figure*}[t]
    \centering
    \includegraphics[width=0.99\textwidth,keepaspectratio=true]{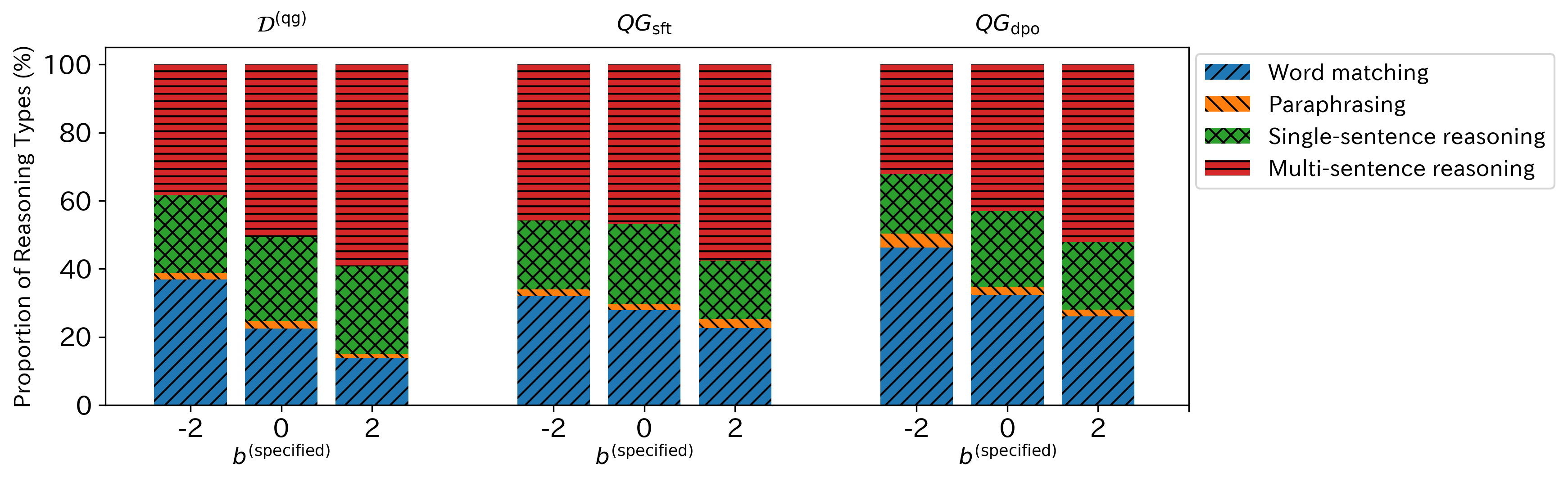}
    \caption{Reasoning types of questions generated by $QG_{\mathrm{sft}}$ and $QG_{\mathrm{dpo}}$ and $\mathcal{D}^{(\mathrm{qg})}$ at each specified difficulty level.}
    \label{fig:reasoning-type}
\end{figure*}

To investigate the relationship between question difficulty and reasoning types, we conducted the following experiment.
\begin{enumerate}
    \item From the QG model training dataset $\mathcal{D}^{(\mathrm{qg})}$, we selected questions with difficulty levels of $-2.0$, $0.0$, or $2.0$, retaining only solvable questions.
    \item We collected all questions generated by $QG_{\mathrm{sft}}$ and $QG_{\mathrm{dpo}}$ in Section~\ref{sec:eval-process} at specified difficulty levels $b^{(\mathrm{specified})} = -2.0, 0.0, 2.0$, again retaining only solvable questions.
    \item We used GPT-4.1 mini to classify each question into one of the four reasoning types.
          The classification prompt is provided in Appendix~\ref{appendix:eval-prompt}.
\end{enumerate}

Figure~\ref{fig:reasoning-type} shows the results, with the horizontal axis representing the difficulty level, and the vertical axis indicating the proportion of reasoning types.
The figure shows that the proportion of word matching questions decreases as the difficulty level increases, whereas the proportion of multi-sentence reasoning questions increases. This trend implies a positive relationship between question difficulty and reasoning type.
Furthermore, it can be observed that this trend is more pronounced in $QG_{\mathrm{dpo}}$ than in $QG_{\mathrm{sft}}$, providing further evidence that $QG_{\mathrm{dpo}}$ achieves more accurate difficulty control.


\section{Conclusion}

In this study, we proposed a difficulty-controllable multiple-choice QG method for reading comprehension, leveraging an LLM with a training method based on the DPO technique to improve the accuracy of difficulty control.
Evaluation using the RACE dataset demonstrated that the proposed method successfully controlled the difficulty of generated multiple-choice questions.
Furthermore, the experiments also showed the effectiveness of our DPO in improving difficulty controllability, while maintaining fluency, content relevance, and answerability.
Additionally, comparison between the proposed method and few-shot learning indicated that few-shot learning is insufficient for achieving difficulty control, highlighting the advantages of the proposed method.

Although our study demonstrates the effectiveness of the proposed QG framework, it has several limitations that warrant further investigation.

The first limitation concerns answerability. As discussed in Section~\ref{sec:eval-others}, the answerability of questions generated by our model remains suboptimal.
A promising direction is to leverage external LLMs to evaluate the generated questions and provide feedback, which can then be incorporated into training as a reward signal through DPO.
We plan to explore this approach with the aim of improving the reliability and usability of generated questions.

The second limitation lies in the reliance on a large-scale training dataset.
Our experiments used the RACE dataset, which contains 87866 questions.
However, assembling such a large dataset is rarely feasible in practical educational settings.
Data augmentation represents a potential solution for mitigating data scarcity.
We intend to investigate data augmentation techniques that enable controllable QG even under low-resource conditions.

The third limitation concerns the impact of the improved accuracy of difficulty estimation achieved by our method. Prior work~\cite{tomikawa2024adaptive} has demonstrated that difficulty control enhances the efficiency of learner ability estimation in adaptive QG, wherein questions are dynamically generated so that their difficulty matches a learner's estimated ability. Future studies will need to evaluate the educational benefits of our approach under such adaptive QG settings.

\appendices
\section{Prompt Design for the Evaluation}
\label{appendix:eval-prompt}

The prompts used for evaluating question quality in Section~\ref{sec:eval-others} are summarized below.
Table~\ref{tb:prompt-fluency} presents the prompt for fluency evaluation,
Table~\ref{tb:prompt-relevance} presents the prompt for relevance evaluation,
and Table~\ref{tb:prompt-answerability} presents the prompt for answerability evaluation.
Furthermore, Table~\ref{tb:prompt-reasoning} presents the prompt for reasoning type classification of questions in Section~\ref{sec:eval-reasoning-type}.

\begin{table}[t]
    \caption{Prompt used for fluency evaluation}
    \label{tb:prompt-fluency}
    \centering
    \small
    \resizebox{\columnwidth}{!}{
        \begin{tabular}{p{\columnwidth}} \hline
            \vspace{0.5pt}
            The following are the text to be comprehended, the corresponding question, and the options. Please evaluate whether the question is fluent to the reading passage on a scale of three: appropriate, acceptable or inappropriate. If appropriate, output ``2''; acceptable, output ``1''; if inappropriate, output ``0''.                            \\
            \#\#\# Context: \{$\bm{c}$\}                                                                                                                                                                                                                                                                                                                        \\
            \#\#\# Question: \{$\bm{q}$\}                                                                                                                                                                                                                                                                                                                       \\
            \#\#\# Options:
            A. \{$\bm{d}_1$\}
            B. \{$\bm{d}_2$\}
            C. \{$\bm{d}_3$\}
            D. \{$\bm{d}_4$\}                                                                                                                                                                                                                                                                                                                    \vspace{0.5pt} \\ \hline
        \end{tabular}
    }
\end{table}

\begin{table}[t]
    \caption{Prompt used for Relevance evaluation}
    \label{tb:prompt-relevance}
    \centering
    \small
    \begin{tabular}{p{\columnwidth}} \hline
        \vspace{0.5pt}
        The following are the text to be comprehended, the corresponding question, and the options. Please evaluate whether the question is relevant to the reading passage on a scale of two: appropriate or inappropriate. If appropriate, output  ``1'', and if inappropriate, output ``0''.                \\
        \#\#\# Context: \{$\bm{c}$\}                                                                                                                                                                                                                                                                           \\
        \#\#\# Question: \{$\bm{q}$\}                                                                                                                                                                                                                                                                          \\
        \#\#\# Options:
        A. \{$\bm{d}_1$\}
        B. \{$\bm{d}_2$\}
        C. \{$\bm{d}_3$\}
        D. \{$\bm{d}_4$\}                                                                                                                                                                                                                                                                       \vspace{0.5pt} \\

        \hline
    \end{tabular}
\end{table}

\begin{table}[t]
    \caption{Prompt used for answerability evaluation}
    \label{tb:prompt-answerability}
    \centering
    \small
    \resizebox{\columnwidth}{!}{
        \begin{tabular}{p{\columnwidth}} \hline
            The following are the text to be comprehended, the corresponding question, and the options. If there is only one correct answer among the options, output ``1''. If there are two or more correct answers or no correct answer at all, output ``0''. Additionally, provide the reason. \\
            \#\#\# Context: \{$\bm{c}$\}                                                                                                                                                                                                                                                           \\
            \#\#\# Question: \{$\bm{q}$\}                                                                                                                                                                                                                                                          \\
            \#\#\# Options:
            A. \{$\bm{d}_1$\}
            B. \{$\bm{d}_2$\}
            C. \{$\bm{d}_3$\}
            D. \{$\bm{d}_4$\}                                                                                                                                                                                                                                                                      \\ \hline
        \end{tabular}
    }
\end{table}

\begin{table}[t]
    \caption{Prompt used for reasoning type classification}
    \label{tb:prompt-reasoning}
    \centering
    \resizebox{\columnwidth}{!}{
        \begin{tabular}{p{\columnwidth}} \hline
            Read the provided definitions carefully and determine how the following question relates to the given context. Classify the relationship by selecting exactly one category from among the provided options. Your response should contain only the exact category name, without any additional explanation.

            \#\#\# Definitions:                                                                                                                                      \\
            Word matching: The question exactly matches a span in the article. The answer is self-evident.                                                           \\
            Paraphrasing: The question is entailed or paraphrased by exactly one sentence in the passage. The answer can be extracted within the sentence.           \\
            Single-sentence reasoning: The answer can be inferred from a single sentence of the article by recognizing incomplete information or conceptual overlap. \\
            Multi-sentence reasoning: The answer must be inferred by synthesizing information distributed across multiple sentences.                                 \\
            \#\#\# Context: \{$\bm{c}$\}                                                                                                                             \\
            \#\#\# Question: \{$\bm{q}$\}                                                                                                                            \\
            \#\#\# Options:
            A. \{$\bm{d}_1$\}
            B. \{$\bm{d}_2$\}
            C. \{$\bm{d}_3$\}
            D. \{$\bm{d}_4$\}                                                                                                                                        \\ \hline
        \end{tabular}
    }
\end{table}

\bibliographystyle{IEEEtran}
\bibliography{paper}

\begin{IEEEbiography}[
        {\includegraphics[width=1in,height=1.25in,clip,keepaspectratio]{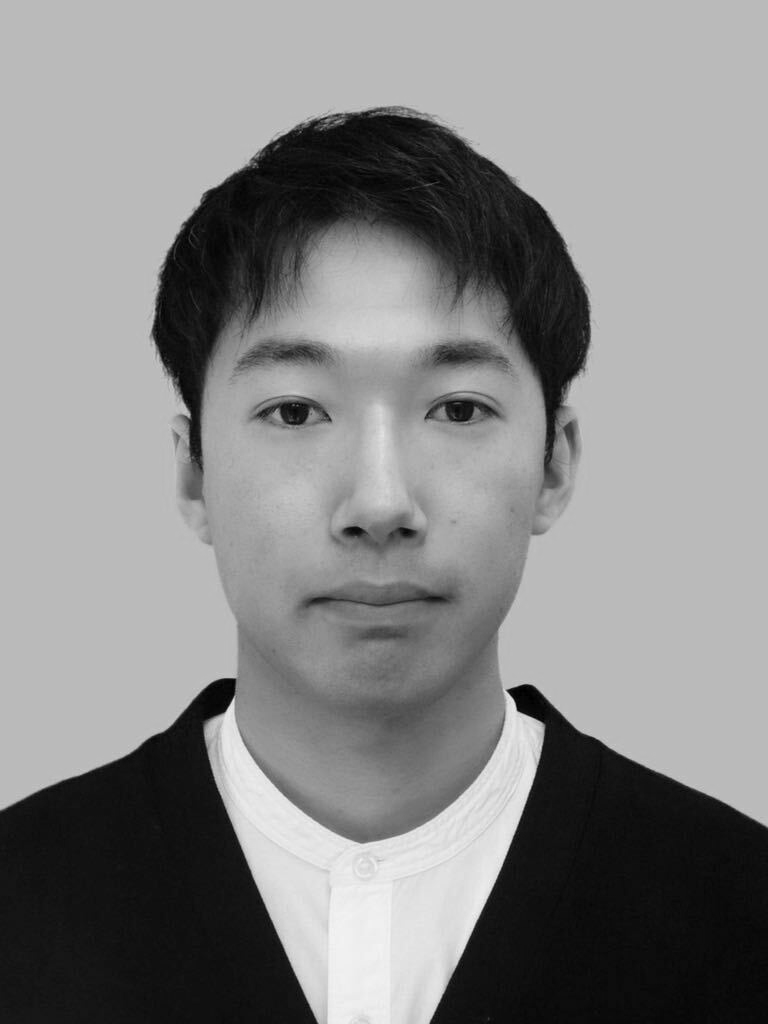}}]{Yuto Tomikawa}
    received his M.S. in Engineering from the University of Electro-Communications, Chofu, Tokyo, Japan in 2025, where he is currently working on a Ph.D. in Engineering.
    His research interests include educational and psychological measurement, machine learning, and natural language processing.
\end{IEEEbiography}
\begin{IEEEbiography}[
        {\includegraphics[width=1in,height=1.25in,clip,keepaspectratio]{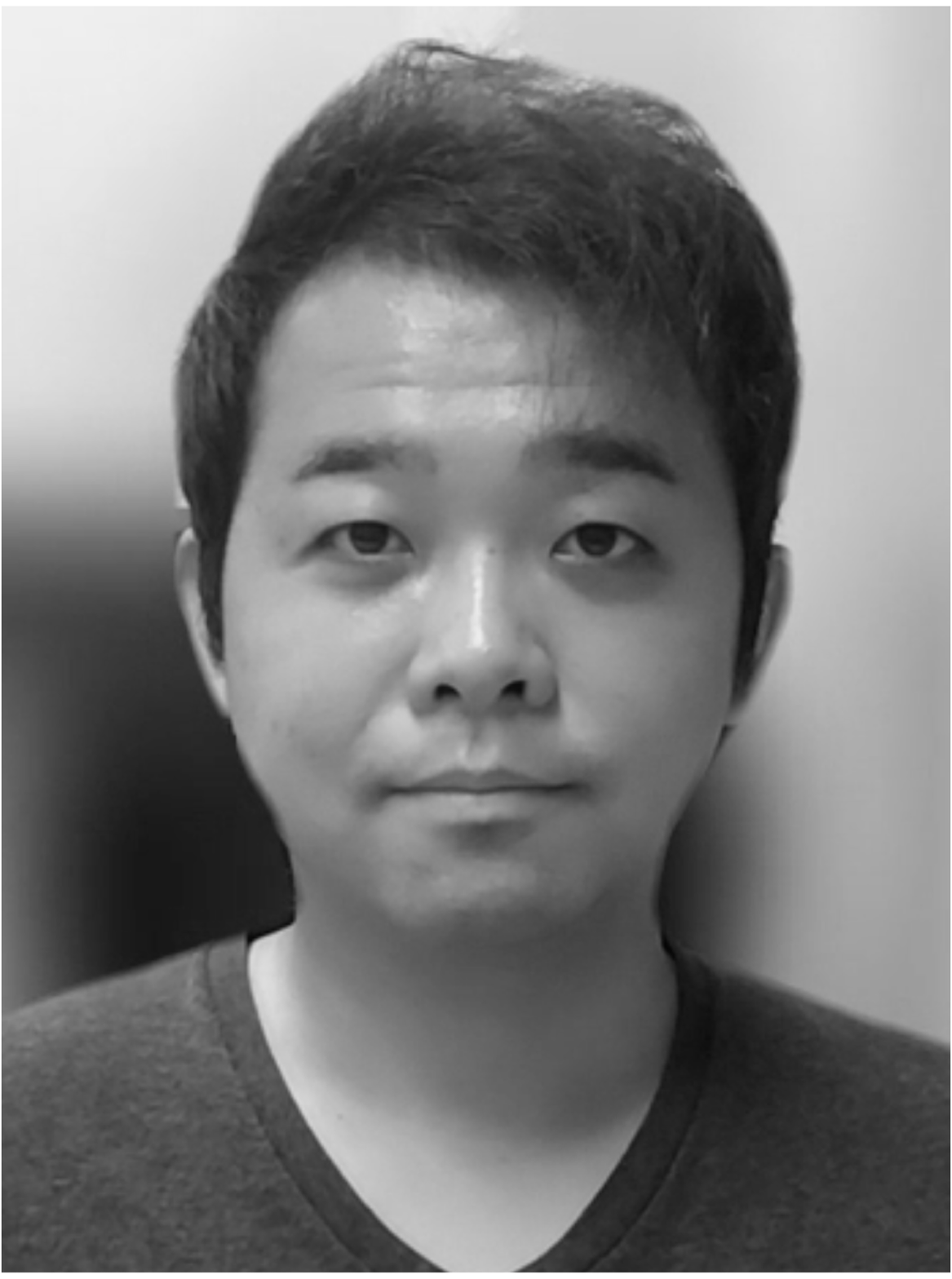}}]{Masaki Uto}
    received his Ph.D. in Engineering from the University of Electro-Communications, Chofu, Tokyo, Japan in 2013.
    He has been an Associate Professor at the Graduate School of Informatics and Engineering, University of Electro-Communications, since 2020.
    His research interests include educational and psychological measurement, Bayesian statistics, machine learning, and natural language processing.
    Dr. Uto was the recipient of the Best Paper Runner-Up Award at the 2020 International Conference on Artificial Intelligence in Education.
\end{IEEEbiography}

\end{document}